\documentclass[runningheads]{llncs}
\usepackage{graphicx}
\usepackage{lineno}
\usepackage{color}
\usepackage{amsmath}
\usepackage{amssymb}
\usepackage{amsfonts}

\def\cocoa{{\hbox{\rm C\kern-.13em o\kern-.07em C\kern-.13em o\kern-.15em A}}}

\begin{document}
\renewcommand\thelinenumber{\color[rgb]{0.2,0.5,0.8}\normalfont\sffamily\scriptsize\arabic{linenumber}\color[rgb]{0,0,0}}
\renewcommand\makeLineNumber {\hss\thelinenumber\ \hspace{6mm} \rlap{\hskip\textwidth\ \hspace{6.5mm}\thelinenumber}}
\pagestyle{headings}
\mainmatter

\title{The Five Points Pose Problem : A New and Accurate Solution Adapted to any
Geometric Configuration}

\author{\textbf{Mahzad Kalantari \textsuperscript{a,b,c}, Franck Jung\textsuperscript{d}, Jean-Pierre Guedon\textsuperscript{b,c} and Nicolas Paparoditis\textsuperscript{a}}}
\institute{\textsuperscript{a}MATIS Laboratory, Institut G´eographique National\\
 2, Avenue Pasteur. 94165 Saint-Mand\'e Cedex - FRANCE - 
firstname.lastname@ign.fr\\
\textsuperscript{b} Institut Recherche Communications Cybern\'etique de Nantes (IRCCyN) UMR CNRS 6597\\ 
 1, rue de la No\"{e} BP 92101F-44321 Nantes Cedex 03 - FRANCE \\
\textsuperscript{c} Institut de Recherche sur les Sciences et Techniques de la Ville CNRS FR 2488\\
\textsuperscript{d} DDE - Seine Maritime - 
firstname.lastname@equipement.gouv.fr\\}

\maketitle

\begin{abstract}
The goal of this paper is to estimate directly the rotation and translation between two stereoscopic images with the help of five homologous points. The methodology presented does not mix the rotation and translation parameters, which is comparably an important advantage over the methods using the well-known essential matrix. This results in correct behavior and accuracy for situations otherwise known as quite unfavorable, such as planar scenes, or panoramic sets of images (with a null base length), while providing quite comparable results for more "standard" cases.
The resolution of the algebraic polynomials resulting from the modeling of the coplanarity constraint is made with the help of powerful algebraic solver tools (the Gr\"{o}bner bases and the Rational Univariate Representation).

\end{abstract}
\section{Introduction}
The determination of the relative orientation between two cameras with the help of homologous points is the basis of many applications in the domains of photogrammetry and computer vision.
The configuration often called "minimal case problem" takes the intrinsic parameters (i. e. the elements of calibration) of the camera as a priori known. Then only five points homologous are necessary to estimate the remaining three unknowns of rotation and two ones of translation (up to a scale factor). This problem has been dealt by many authors, and most of recent methods published provide a resolution based on the properties of the essential matrix. Even if its use simplifies remarkably the problem of the relative orientation, in some cases, due to the fact that all parameters of rotation and translation are mixed, this is the origin of geometric ambiguousnesses. So as to improve this point, we propose in this article a model that separates completely the rotation and the translation unknowns. We show that the major advantage of this model is that it allows to solve degenerate problems such as pure rotations (null translation). We use an algebraic modeling for the coplanarity constraint, through a system of polynomial equations. We solve them with the help of powerful algebraic solver tools, the Gr\"{o}bner bases and the Rational Univariate Representation.
So as to assess this new approach, three cases have been processed: a classical case, a planar scene, and a case where the base length is close to zero. We will see that the new method is still accurate even for the last two cases - quite unfavorable - configurations. We will also compare with the Stewenius's algorithm and see that in planar scenes the new algorithm is more accurate. An evaluation on real scenes will finally be presented.
\section{Historical background of the five points relative pose problem.}
It was for the first time demonstrated by Kruppa \cite{KruppaEq} in 1913 that the direct resolution of the relative orientation from 5 points in general contained at most 11 solutions. The described method consisted to find all intersections of two curves of degree 6. Unfortunately, one century ago, this method could not lead to a numerical implementation. Lately in \cite{Demazure}, \cite{FaugerasMIT}, \cite{FaugerasMaybank}, \cite{HeydenReconsDemazure} it has been demonstrated that the number of solutions is in general equal to 10, including the complex solutions. Triggs \cite{Triggs5points} has provided a detailed version for a numeric implementation. Philip \cite{Philip5points} presented in 1996 a solution using a polynomial of degree 13, and has proposed a numeric method to solve his system. The roots of his polynomial give directly the relative orientation. Philip has exploited the constraints on essential matrix. Philip's ideas have been followed in 2004 by Nister \cite{Nister04} who has refined this algorithm, has obtained a $10^{th}$ order polynomial and has given a numerical resolution using a Gauss-Jordan elimination. Since then, number of papers tried to give some improvements to the method of Nister, notably Stewenius \cite{stewenius-engels-etal-ijprs-06} that has provided a polynomial resolution using the Gr\"{o}bner bases. Many papers have proposed some modifications to the method of Nister in view of a numeric improvement \cite{BatraFiveAlernative}, \cite{PlanarProblemNister}, or for a simplification of implementation \cite{LiHartleyfiveEasy}, \cite{Sarkis_ICASSP07}.
\section{Geometry review of Relative Orientation}
\begin{figure}[!h]
\centering
\includegraphics[height=3cm]{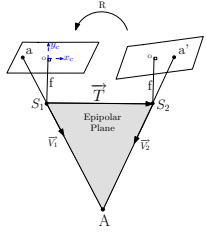}
\caption{Geometry of relative orientation}
\label{fig:copla}
\end{figure}

In this section we recall the various ways to present the geometry of relative orientation, that consist in the determination of the translation and the relative rotation between two images of a scene having a common informational part. 
In general, the position of the first camera is taken as the origin of the system $S_1$ (Fig. \ref{fig:copla}) and therefore the position of the second camera ($S_2$) is calculated in relationship to the first one. $O$ named as the principal point of the camera, $f$ is the focal length. $A$ is the world point, and the image projection of A on the left image is $a$ with  coordinates $(x_a, y_a, f)^T$, and $a'$ $(x_{a'}, y_{a'}, f)^T$ in the right image. The vector of translation $\overrightarrow{T}$ $(T_x,T_y,T_z) $ is the basis that relies the optic centers of the cameras ($S_1$ and $S_2$). $R$ is the relative rotation between the two cameras.
A way for modelling the relative orientation is known as condition of coplanarity. This constraint has been often used by the community of computer vision since three decades. As pictured in the Fig. 1, the condition of coplanarity between two images expresses the fact that the vector $\overrightarrow{V_1}$, the vector $\overrightarrow{V_2}$(expressed in the reference of $\overrightarrow{V_1}$), and the vector of the translation  $\overrightarrow{T}$ are in the same plane, called the epipolar plane. One can translate this condition by a null value for the triple product between these 3 vectors. In other words:
\begin{equation}\label{produitMixte}
\overrightarrow{V_2} \cdot (R \overrightarrow{V_1} \wedge \overrightarrow{T})=0
\end{equation}

\section{Algebraic modelling of the five-points problem}
In this section, different ways for algebraic modelling of the relative orientation are recalled. The goal is to obtain a polynomial system, so as to use the powerful mathematical tools developed for solving such systems.\\
The coplanarity constraint (equation \ref{produitMixte}) under its algebraic shape is expressed by the equation:
\begin{equation}\label{algebricCoplanarity}
\begin{bmatrix}x_{a'}&y_{a'}&f\end{bmatrix}\begin{bmatrix}\,0&\!T_z&\,\,\,-T_y\\\,\,\,-T_z&\,0&\!T_x\\T_y&\,\,-T_x&\,0\end{bmatrix}\begin{bmatrix}\,r_{11}&\!r_{12}&\,\,\,r_{13}\\\,\,\,r_{21}&\,r_{22}&\!r_{23}\\r_{31}&\,\,r_{32}&\,r_{33}\end{bmatrix}\begin{bmatrix}x_{a}\\y_{a}\\f\end{bmatrix}=0.
\end{equation}
In this equation the unknowns are the matrix of rotation $R$ and the translation $T$.
Different ways exist to parameter the system so as to obtain polynomials with the rotation and the translation as unknowns.
In the present part the main modelling solutions for the rotation and the translation to be used in this research are described.
\subsection{Modelling of the translation}
The translation of unity length between the two centres of the cameras may be understood as imaging on the unity sphere its center. The translation has only 2 degree of freedom, and for that reason, with the relative orientation, the scale cannot be determined.
The equation of the unity sphere is the following :
\begin{equation} \label{equationTranslation}
T_x^2+T_y^2+T_z^2 =1.
\end{equation}
The advantage of this constraint of normality on the translation is that it is quite possible to work even with a very small translation, allowing to compute precisely the rotation when the translation is null. In this case, as the base $\overrightarrow{T}$ is null, the two homologous vectors $\overrightarrow{V_1}$ and $\overrightarrow{V_2}$ are deduced from each other by a rotation $R$, so that $R\overrightarrow{V_1}\wedge \overrightarrow {V_2} = \overrightarrow{0}$. Thus the triple product $\overrightarrow{T} \cdot (R \overrightarrow{V_1} \wedge \overrightarrow{V_2})$ is null whatever $\overrightarrow{T}$. 
The fact that we force the translation to be unity prevents that $\overrightarrow{T} = \overrightarrow{0}$ and therefore to suffer numeric instabilities. This implies in turn that the rotation will be correctly estimated in any case.
\subsection{Modelling of the Rotation in 3D space}
The rotation matrix $(R)$ in the 3D space has 3 degree of freedom. It is thus possible to express it with 3 parameters. However several representations with more than 3 parameters exist. The algebraic model will depend on the chosen representation. In the following part the main models for the coplanarity constraint are described.
\subsubsection{Representation of the rotation using quaternions (4 parameters)}
A quaternion is composed of 4 parameters, $q=(a,b,c,d)^t$, with the vector part being $(b, c, d)$. The quaternions provide a simple representation of the rotation. Indeed with the parameters of a unity quaternion , the matrix of rotation can be expressed in the following manner :
\begin{equation}\label{quaternionMatrix}
\left[ \begin {array}{ccc} 1-2\,{c}^{2}-2\,{d}^{2}&2\,bc-2\,da&2\,bd+2\,ca\\\noalign{\medskip}2\,bc+2\,da&1-2\,{b}^{2}-2\,{d}^{2}&2\,cd-2\,
ba\\\noalign{\medskip}2\,bd-2\,ca&2\,cd+2\,ba&1-2\,{b}^{2}-2\,{c}^{2}
\end {array} \right]
\end{equation}
With such a model the number of unknowns for the rotation is 4.

\subsubsection{Representation using Thompson rotation.}\label{Polynome7}
Another efficient way to represent the rotation with three parameters is given in Thompson's paper \cite{ThompsonRota}.
\begin{equation}\label{Thomson}
\frac{1}{\Delta} \left[
  \begin{array}{ c c c}
     {\Delta}' & -\nu & \mu \\
    \nu &  {\Delta}' & -\lambda\\
     -\mu  & \lambda & {\Delta}'
  \end{array} \right] +
 \frac{1}{2\Delta}\left[
  \begin{array}{ c}
     \lambda  \\
     \mu \\
     \nu
  \end{array} \right] \left[
  \begin{array}{ c c c }
     \lambda & \mu & \nu\\
  \end{array} \right]\\
\end{equation}
where ${\Delta}= 1+\frac{1}{4}(\lambda^2+\mu^2+\nu^2)$ and ${\Delta}'= 1- \frac{1}{4}(\lambda^2+\mu^2+\nu^2).$
With such a model the number of unknowns for the rotation also resumes to three.
Other models of rotation matrix exist, such as the Cayley transfom, often used in robotics \cite{RouillierRobotique}.
\subsection{Algebraic modelling of the coplanarity constraint}
\subsubsection{Model with 7 equations}
With the model of rotation matrix given by equation \ref{quaternionMatrix}, every couple of homologous points ($a$ and $a'$) gives an equation of the type :
\begin{multline}\label{quaternionPolynome7}
({ xa'}_{{i}} ( -{ T_z} ( 2bc+2da ) +{ T_y} ( 2bd-2ca )  ) +{ ya'}_{{i}}( { T_z} ( 1-2{c}^{2}-2{d}^{2} )-
{ T_x}( 2bd-2ca ))+\\
{ za'}_{{i}} ( -{ T_y}( 1-2{c}^{2}-2{d}^{2} ) +{ T_x} ( 2bc+2da )  )  ) { xa}_{{2}}+ ( { xa'}_{{i}}(-{ T_z} ( 1-2{b}^{2}-2{d}^{2} ) +{ T_y}( 2cd+2ba))+\\{ ya'}_{{i}} ( { T_z}( 2bc-2da ) -{ T_x} ( 2cd+2ba ) )+{ za'}_{{i}} ( -{ T_y} ( 2bc-2da) +{ T_x} ( 1-2{b}^{2}-2{d}^{2}))){ ya}_{{2}}+ \\( { xa'}_{{i}} ( -{T_z}(2cd-2ba ) +{ T_y} ( 1-2{b}^{2}-2{c}^{2})) +{ ya'}_{{i}} ( { T_z} ( 2bd+2ca) -{ T_x} ( 1-2{b}^{2}-2{c}^{2} )) +\\{ za'}_{{i}} ( -{ T_y} ( 2bd+2ca ) +{T_x} ( 2cd-2ba ))) { za}_{{2}}=0
\end{multline}
5 couples of points gives 5 equations. It is otherwise necessary to add along with this equations the constraint of normality on the quaternion, as well as on the translation.
\begin{eqnarray}
\label{contraintesquaternion}
 T_x^2+T_y^2+T_z^2 = 1,\\
 a^2+b^2+c^2+d^2 =1.
\end{eqnarray}
Finally, the unknowns of the system are $[a, b, c, d, T_x, T_y, T_z]$
\subsubsection{Model with 6 equations}  \label{Polynome6}
While using the Thompson rotation matrix, the rotation is expressed with 3 parameters. The system will have 6 unknowns, considering the three parameters of translation.
The polynomial expressing the coplanarity constraint for a couple of homologous points, taking for model the Thompson rotation, is the following :
\begin{multline} \label{quaternionsPolynome6thom}
({xa}_{{i}} ( -4{ T_z}{\nu}-2{ T_z}{\lambda}{\mu}-4{ T_y}{\mu}+2{ T_y}{\lambda}{\nu} ) +{ya}_{{i}} (4{T_z}+{T_z}{{\lambda}}^{2}-{ T_z}{{\mu}}^{2}-{ T_z}{{\nu}}^{2}+\\4{ T_x}{\mu}-
2{ T_x}{\lambda}{\nu} ) +{ za}_{{i}} ( -4{ T_y}-{ T_y}{{\lambda}}^{2}+{ T_y}{{\mu}}^{2}+
{ T_y}{{\nu}}^{2}+4{ T_x}{\nu}+2{ T_x}{\lambda}{\mu})) { xa'}_{{i}}+ \\( { xa}_{{i}} ( -4
{ T_z}+{ T_z}{{\lambda}}^{2}-{ T_z}{{\mu}}^{2}+{ T_z}{{\nu}}^{2}+4{T_y}{\lambda}+
2{ T_y}{\mu}{\nu} ) +{ ya}_{{i}} ( -4{ T_z}{\nu}+2{ T_z}{\lambda}{\mu}-4{ T_x}{\lambda}-\\2{ T_x
}{\mu}{\nu} ) +{ za}_{{i}} ( 4{ T_y}{\nu}-2{ T_y}{\lambda}{\mu}+4{ T_x}-
{ T_x}{{\lambda}}^{2}+{ T_x}{{\mu}}^{2}-{ T_x}{{\nu}}^{2} )  ) { ya'}_{{i}}+
( { xa}_{{i}} ( 4{ T_z}{\lambda}-2{ T_z}{\mu}{\nu}\\+4{T_y}-{ T_y}{{\lambda}}^{2}-
{ T_y}{{\mu}}^{2}+{ T_y}{{\nu}}^{2} ) +{ ya}_{{i}} ( 4{ T_z}{\mu}+
2{ T_z}{\lambda}{\nu}-4{ T_x}+{ T_x}{{\lambda}}^{2}+{ T_x}{{\mu}}^{2}-
{ T_x}{{\nu}}^{2} ) +\\
{ za}_{{i}} ( -4{ T_y}{\mu}-2{ T_y}{\lambda}{\nu}-
4{ T_x}{\lambda}+2{ T_x}{\mu}{\nu} )  ) { za'}_{{i}}=0
\end{multline}
The constraint of normality on the translation (equation \ref{equationTranslation}) is added to these 5 equations. So the system has 6 equations and 6 unknowns $[\lambda, \mu, \nu, T_x, T_y, T_z]$.

In conclusion of this section, we have built two polynomial systems, where the translation and rotation parameters are distinct and correspond to separated unknowns. 
Next, we show how to solve this type of polynomial systems.

\section{Resolution of the polynomial systems}
The ways to solve the polynomial systems are widely published \cite{CoxGeometry}, \cite{CoxIdeal}, and are briefly recalled for the reader not familiar with this topic. 
The resolution of a polynomial system consists in finding the zeros of an algebraic equation system such as :
$P(x) = 0$ with $P=(p_1,p_2,..,p_n)$ where $p_i$ is a $l-variable$ polynomial $x = (x_1,x_2,...,x_l)$ over the field $\mathbb{C}\ $ of complex numbers.
Differents types of solvers for polynomial equations exist, such as analytic solvers, subdivision solvers, geometric solvers, homotopic solvers and algebraic solvers \cite{ElkadiMourrainPolynome}.
In this paper the focus is on algebraic solvers, that exploit the known relationships between the unknowns. They subdivide the problem of the resolution into two steps : the first consists in transforming the system into one or several equivalent systems, but better adapted, and this constitutes what one will call an algebraic solution. The second step consists, in the case where one works in one subfield of the complex field, to calculate the numeric values of the solutions from the algebraic solution.
We will see now briefly the  principal tools used in this paper for solving polynomial systems.
But first, some remainders of geometric algebra are necessary.
\subsection{Notations}

$\mathbb{Q}\ [X_1,X_2,...,X_n]$ is the polynomial rings with rational coefficients and unknowns $X_1, X_2,..., X_n$. $S = P_1, P_2, ... P_s$ is any subset of $\mathbb{Q}\ [X_1, X_2, ..., X_n]$. A point $x\in \mathbb{C }^n$ is a zero of $S$ if $P_i(x) = 0$ $\forall i = 1, 2,..., s$. The variety of $P$ is the set of all common complex zeros :
\begin{equation}
\mathcal{V}(P) =\{ (a_1, ..., a_n) \in  \mathbb{C}^n: p_i(a_1, ..., a_n) = 0\ for\ all\  1 \leq i \leq s \}   .
\end{equation}
The ideal $\mathcal{I}$ generated by a finite set of multivariate polynomials $<P_1, P_2,..., P_s>$ is defined as:
\begin{equation}
\mathcal{I} = \{\sum_{i=1}^{n}h_i P_i|h_i \in \mathbb{Q}\ [X_1, X_2,..., X_n]\}.
\end{equation}

The ideal contains all polynomials which can be generated as an algebraic combination of its generators. An ideal can be generated by many different sets of generators, which all have the same solutions.
\subsection{Construction of the algebraic solver : an introduction to the Gr\"{o}bner bases.}
A Gr\"{o}bner basis is a set of multivariate polynomials that has "nice" algorithmic properties. Every set of polynomials can be transformed into a Gr\"{o}bner basis. This process generalizes three familiar techniques : the Gauss elimination for solving linear systems of equations, the Euclidean algorithm for computing the greatest common divisor of two univariate polynomials, and the Simplex Algorithm for linear programming [3].
The Gr\"{o}bner bases were developed initially by B. Buchberger in the years 1960 \cite{Buchberger2}.
The first step, when we want to compute a Gr\"{o}bner basis, is to define an \textit{monomial order}. For polynomial rings with severable variables, there are many possible choices of monomial orders. Some of the most important orders are given in the following definitions.
\newtheorem{def1}{Definition}
\begin{def1}
\bf{Lexicographic Order (Lex)}\\
$ X_1^{\alpha_1}\ ...\ X_n^{\alpha_n}\ <_{Lex}\ X_1^{\beta_1}\ ...\ X_n^{\beta_n}\ \Leftrightarrow\ \exists\ i_0\leq\ n\ ,
\left\{
    \begin{array}{ll}
         \alpha_i=\beta_i,\  for\  i = 1,2,...,i_{0-1}\\
         \alpha_{i0}\ < \ \beta_{i0}
    \end{array}
\right.$
\end{def1}
\newtheorem{def3}[def1]{Definition}
\begin{def3}
\bf{Degree Reverse Lexicographic Order (DRL)}\\
$ X_1^{\alpha_1}\ ...\ X_n^{\alpha_n}\ <_{DRL}\ X_1^{\beta_1}\ ...\ X_n^{\beta_n}\ \Leftrightarrow\ X_0^{\sum_{k}^{\alpha_{k}}}X_1^{-\alpha_n}...X_n^{-\alpha_1}\ <_{Lex}\ X_0^{\sum_{k}^{\beta_{k}}}X_1^{-\beta_n}...X_n^{-\beta_1}\ $
\end{def3}
The \textit{DRL} ordering, although somewhat nonintuitive, has some interesting computational properties.\\
The following terms and notation are present in the literature of Gr\"{o}bner bases and will be useful later on.
The degree of a polynomial $P$, denoted $DEG(f)$, is the highest degree of the terms in $P$. The leading term of $P$, denoted $LT(P)$, is the term with the highest degree. The leading coefficient of $P$ denoted $LC(P)$ is the coefficient of the leading term in $P$.
Finally Gr\"{o}bner bases can be defined:
\newtheorem{def5}[def1]{Definition}
\begin{def5}
Fix a monomial order $>$ on $\mathbb{Q}\ [X_1,X_2,...,X_n]$, and \\let $\mathcal{I} \subset \mathbb{Q}\ [X_1, X_2,..., X_n]$ be an ideal. A Gr\"{o}bner base for  $\mathcal{I}$ (with respect to $>$) is a finite collection of polynomials $ G = \{g_1,..., g_t\} \subset \mathcal{I} $ with the property that for every nonzero $f\in \mathcal{I}$, $LT(f)$ can be divided by $LT(g_i)$ for some $i$.
\end{def5}
Two principal questions immediately arise from this definition :
\begin{enumerate}
    \item the \textit{existence} of Gr\"{o}bner bases for any ideal $\mathcal{I} $. \\
Hilbert's Basis Theorem says that : Every ideal $\mathcal{I} $ has a Gr\"{o}bner basis $G$. Furthermore, $\mathcal{I} = <g_1,...,g_t>$.\\
\item the issue of \textit{uniqueness} of the Gr\"{o}bner bases.\\
The Buchberger theorem proves that if we fix a term order, then every non-zero ideal $\mathcal{I}$ has an unique \textit{reduced} Gr\"{o}bner basis with respect to this term order.
\end{enumerate}
There are several possible algorithms to effectively compute Gr\"{o}bner bases. The traditional one is Buchberger's algorithm, it has several variants and it is implemented in most general computer algebra systems like Maple, Mathematica, Singular \cite{SINGULAR}, Macaulay2 \cite{M2}, CoCoA \cite{CocoaSystem} and the Salsa Software \cite{SALSASoft}.
In this paper we use the Salsa Software with the F4 algorithm \cite{FaugereF4}. The Faug\`ere F4 algorithm is based on the intensive use of linear algebra methods.
\subsection{Application of Gr\"{o}bner bases for systems solving}
Gr\"{o}bner bases ($G$) give important informations about the initial system of polynomial equations. Here some of these informations with their application on the 2 systems of polynomials defined in Section \ref{Polynome7} and Section \ref{Polynome6} are summarized:
\begin{enumerate}
    \item \textit{Solvability of the polynomial system}.
It is easy to check if it is possible to solve the system of polynomials or not with the help of the Gr\"{o}bner bases $G$. If $G=\{1\}$, the system has no solution. We check this on our two systems, and we find that: $G\neq \{1\}$. In other terms $\mathcal{V}$ is not empty.
\item \textit{Finite solvability of polynomial equations}.
It is easy to see whether the system has a finite number of complex solutions or not : we just check that for each $i$, $ 1 \leq i \leq n $, there is an $m_i \geq 0$ such that $x_{i}^{m_{i}}\ = LT(g)$ for some $g\ \in G$. This type of system is called \textit{zero-dimensional system}. In this case, the set of solutions does not depend on the chosen algebraically closed field.
If we apply this on two systems, we find that the dimension of the two systems is zero. So the set of solutions is finite.
\item \textit{Counting number of finite solutions of the polynomial system}. One important information is that the Gr\"{o}bner basis also gives the number of solutions of the system. If we suppose that the system of polynomial equations $P$ has a finite number of solutions, then the number of solutions is equal to the cardinality of the set of monomials that are not multiple of the leading terms of the polynomials in the Gr\"{o}bner basis (any term ordering may be chosen).
This monomials are called \textit{basis monomials} or \textit{standard monomials} ($B$).\\
For the system of 7 polynomial equations (Section \ref{Polynome7}), the number of the standard bases is equal to 80. It means that the system has a total of 80 complex solutions. This high figure is due to the symmetries of the 2 equations of constraints, for example the quaternion $Q (a, b, c, d)$ is equal to $Q'(-a, -b, -c, -d)$. It has been reminded previously in the literature that the maximum number of solutions for the essential matrix is 10. Every essential matrix gives 4 families of solutions, rotation and translation, which provides 40 families of solutions. In the present system, with the symmetry on the quaternion, there are twice more solutions, therefore the total is 80.\\
Using the system of polynomial equations defined in Section \ref{Polynome6}, the standards bases of this system are the following (in the \textit{DRL order}):
\begin{multline}\label{baseStandard40}
B = [1, T_z, T_y, T_x, \nu, {\mu}, {\lambda} , T_z^2, T_yT_z, T_y^2, T_xT_z, T_xT_y,{\nu}T_z, {\nu}T_y, {\nu}T_x, {\nu}^2, {\mu}T_z,\\
{\mu}T_y, {\mu}T_x, u{\nu}, {\mu}^2, {\lambda}T_z, {\lambda}T_y, {\lambda}T_x, {\lambda}{\nu}, {\lambda}{\mu},{\lambda}^2, T_z^3, T_yT_z^2, T_y^2T_z, T_xT_z^2,\\
T_xT_yT_z, {\nu}T_z^2, {\nu}T_yT_z, {\nu}^2T_z, {\mu}T_z^2, {\mu}T_yT_z, {\mu}{\nu}T_z, {\lambda}T_z^2, {\lambda}{\nu}T_z ]\\
\end{multline}
Which makes a total of 40 bases and therefore 40 solutions.
In the present paper the Salsa library has been used. It is quite possible to use the resolutions according to Stewenius \cite{stewenius-engels-etal-ijprs-06} and \cite{GobnerDistortion}, but it requires a heavier implementation work.
\end{enumerate}
\subsection{Finding the real roots of the polynomial systems}
Once the Gr\"{o}bner basis is calculated, different ways exist to find the roots of the system of polynomial equations, e.g. the method that solves the polynomial systems with the help of elimination and \textit{lex} Gr\"{o}bner basis. Another most popular way to solve polynomial systems is via eigenvalues and eigenvectors, with the help of \textit{standard monomials } \cite{CoxGeometry}. Here  the emphasis is put on the other method, called the Rational Univariate Representation (abbreviated by RUR) \cite{RouRUR}.
Representing the roots of a system of polynomial equations in the RUR was first
introduced by Leopold Kronecker \cite{Kron}, but started to be used in computer algebra only recently \cite{rojas99solving}, \cite{RURGonzalesVega}, \cite{RouRUR}.
 The RUR is the simplest way for representing symbolically the roots of a zero-dimensional system without loosing information (multiplicities or real roots) since one can get all the information on the roots of the system by solving univariate polynomials.
Let $P(X) = 0$ ($<P_1,P_2,...P_s>$) where $P_i\ \in \mathbb{Q}\ [X_1,X_2,...,X_n]$ be a zero-dimensional system with its solution set $\mathcal{V} =P^{-1}(0)$, Rational Univariate Representation of $\mathcal{V}$ consists in expressing all the coordinates as functions of the roots of a univariate polynomial such as :
\begin{equation}\label{RUR}
f_0(T)=0,\ X_1= \frac{f_1(T)}{q(T)}, X_2= \frac{f_2(T)}{q(T)},....,X_n= \frac{f_n(T)}{q(T)}
\end{equation}
where $f_0, f_1, f_2,... , f_n, q \in \mathbb{Q}\ [T]$ ($T$ is a new variable).
Computing a RUR reduces the resolution of a zero-dimensional system to solving one polynomial with one variable $(f_t)$ and to evaluate $n$ rational fractions ($\frac{f_i(T)}{q(T)}, i= 1,...,n$) as its roots. The goal is to compute all the real roots of the system (and only the real roots), providing a numerical approximation with an arbitrary precision (set by the user) of the coordinates. Many efficients algorithms have been implemented to calcultate RUR. More details are easily found in the literature, but a complete explication can be found in \cite{RURTexas}, \cite{RouRUR}, \cite{RouZeroDim}.
An implementation of the Rouillier algorithm for RUR computation can be found in the SALSA software \cite{SALSASoft}.
\section{Algorithm Outlines} \label{section:Outline}
\begin{figure}[!h]
\centering
\includegraphics[scale=0.3]{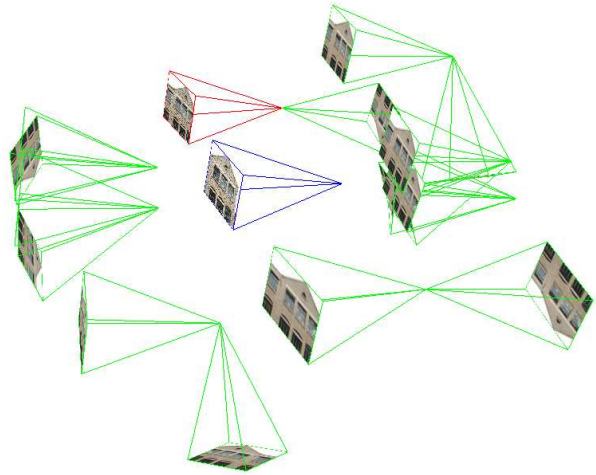}
\caption{Representation of all real solutions}
\label{fig:AllSolution}
\end{figure}

Now, the different steps of our algorithm for the calculation of the relative orientation are described.\\
\textit{Step 1}: 5 couples of homologous points are randomly selected with the RANSAC method \cite{RanSac} \cite{HartleyMultipleView} \cite{YIMA}.\\
\textit{Step 2}: Build the system of polynomial equations.\\
\textit{Step 3}: Solve polynomials system. In the present paper the Salsa library has been used (it is quite possible to use the resolutions according to Stewenius \cite{stewenius-engels-etal-ijprs-06} and \cite{GobnerDistortion}, it requires a heavier implementation work, but has the advantage that a great part of the calculation is made off-line).\\
\textit{Step 4}: Identify the solution with a physical sense. The ambiguity resolution may be done through the use of a third image \cite{Nister04}, but we prefer to be able to work with only two images. It is important to find the "true" solution in this very large set, and it is necessary to inject information bound to the geometry of the scene. We proceed in this way :
\begin{itemize}
    \item when intersecting the rays relative to all homologous couples of points, one keeps the solution where the rays cut themselves in front of the image,
\item for the 5 randomly selected couples, we calculate the distance to the world points. We keep only the solutions that give a depth superior to the value of the normalised baseline, i. e. 1, the other ones being considered as unrealistic,
\item the last step consists in selecting the solution which fits with the highest number of points. This hypothesis requires of course to have more than five points. Other methods exist to find the good solution among all those produced by the direct resolutions, but in general they consist in using a third image \cite{Nister04}.
\end{itemize}
As an example, Fig. \ref{fig:AllSolution} illustrates the different steps. The reference image is in blue frame, all real solutions are presented in green frame, and the final solution has a red frame.
\section{Results and Evaluation}
Here we present the results of an experimentation on both synthetic and real data.
\subsection {Experimentation on Synthetic Data} 
\begin{figure}[!h]
    \centering
    \begin{tabular}{ccc}
        \includegraphics[width=0.35\linewidth]{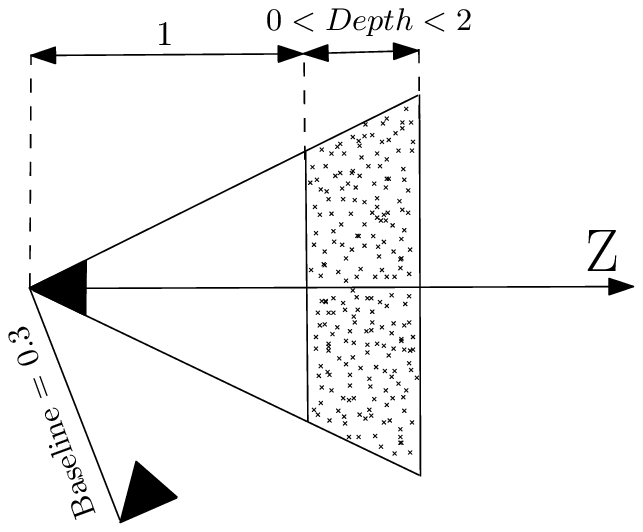}&
        \includegraphics[width=0.25\linewidth]{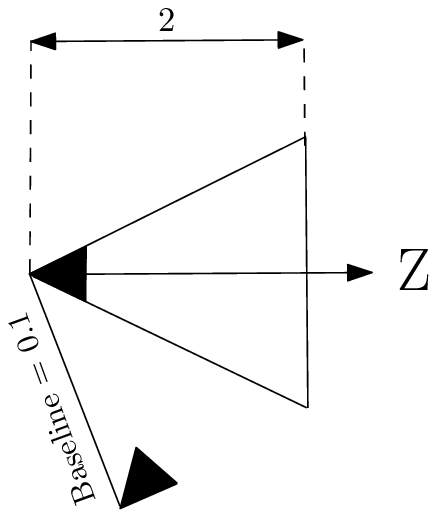}&
        \includegraphics[width=0.35\linewidth]{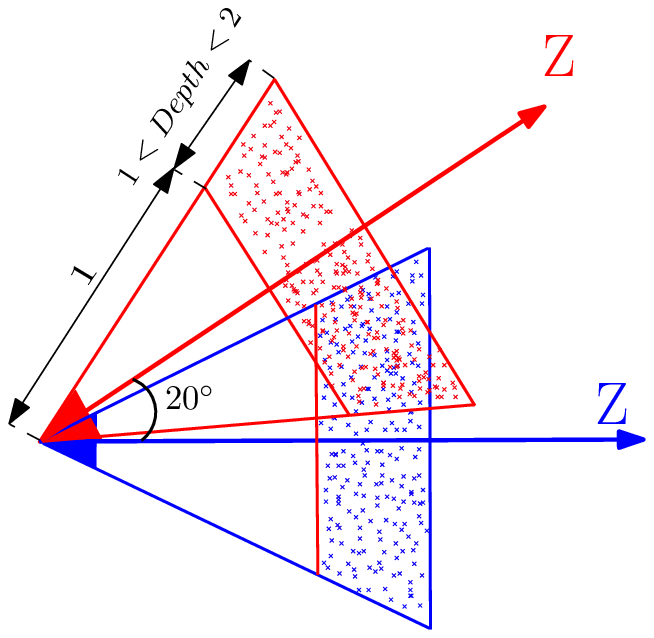}\\
      (a) & (b)&(c)\\
    \end{tabular}
    \caption{(a) Easy condition (b) Planar Condition (c) Zero translational condition
\label{fig:allConfigSimu}
}
\end{figure}

To quantify the performances of the presented method, synthetic data have been simulated. The parameters used for the simulations, are the same as Nister's ones. The images size is 352 x 288 pixels (CIF).
The field of view is 45 degrees wide. The distance to the scene is equal to 1. Several cases have been treated :

\begin{enumerate}
 \item Simple configuration : the baseline between the 2 images has a length of 0.3, the depth varies from 0 to 2.
\item Planar Structure and short baseline (0.1) : a degenerate case where all simulated points are on the plane Z = 2.
\item Zero translation : the configuration of the points is the same as in the simple configuration, the main difference is that the baseline length is null.
\end{enumerate}
In each configuration a Gaussian noise with a standard deviation varying between 0 and 1 pixel is added. The results are average of 100 times independant experiments.
For each situation the minimal case only has been treated, corresponding to the minimum number of points required (5). No least square adjustment has been done. The geometry of the different configurations is illustrated in the Fig. \ref{fig:allConfigSimu}.
\subsubsection{Results in the so-called easy configuration}.
As a comparison our method has been confronted with Stewenius's one, thanks to the codes that he kindly made downloadable on his website \cite{Code5pointsST}. This allowed us to use it as a reference for the algorithm of 5 points. Two sorts of translations have been treated, one in X (sideway motion) and one in Z (forward motion). Our results are mostly similar, even slightly better.
Remark: In these simulations, it is important to specify that the rotation between the two optical axes is always very well determined. The difference between the different methods is the precision of evaluation on the orientation of the base, so this is the assessment that we have used.
\begin{figure}[!h]\label{fig:easy}
    \centering
    \begin{tabular}{cc}
      \includegraphics[width=0.4\linewidth]{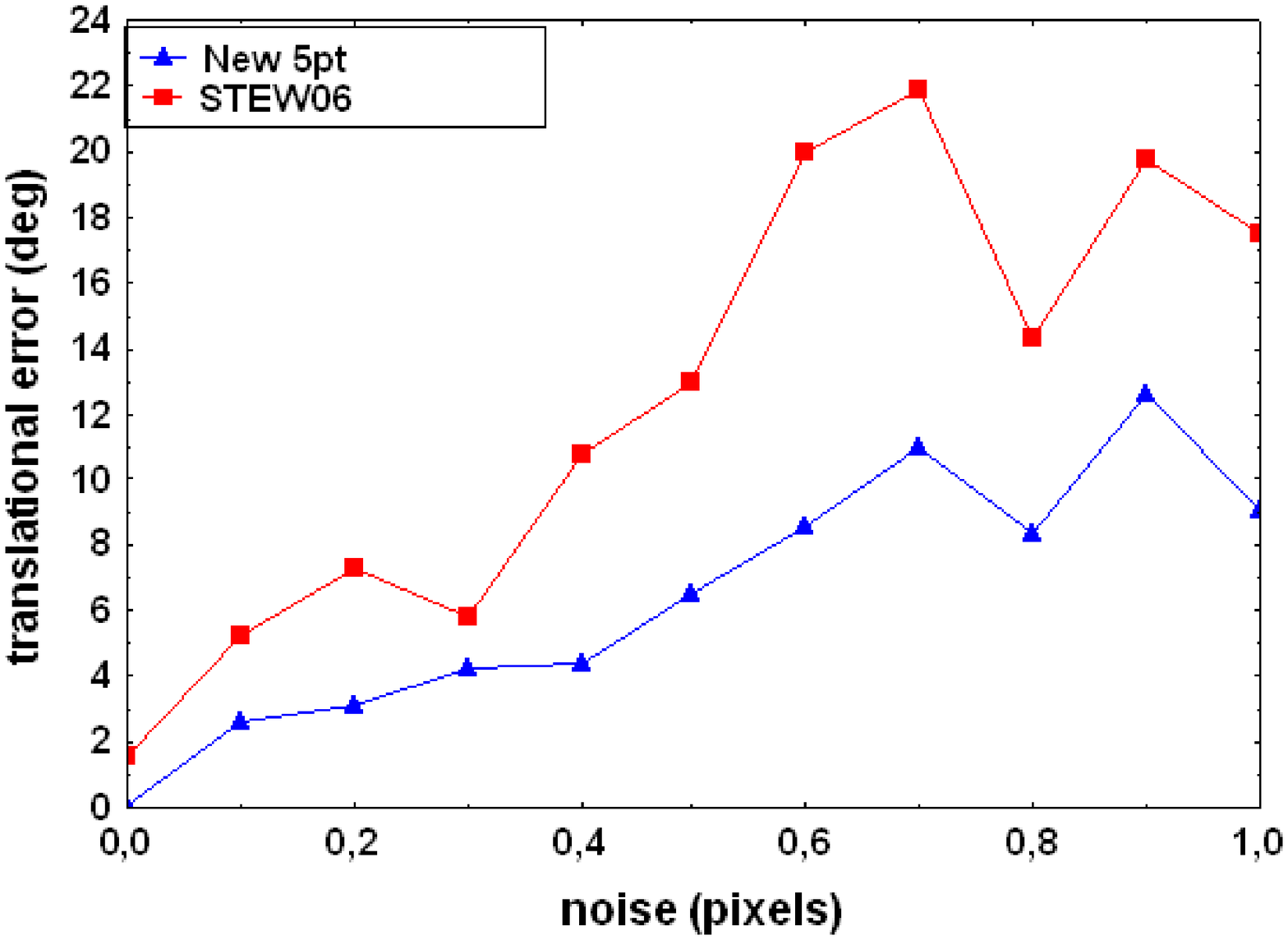} &
      \includegraphics[width=0.4\linewidth]{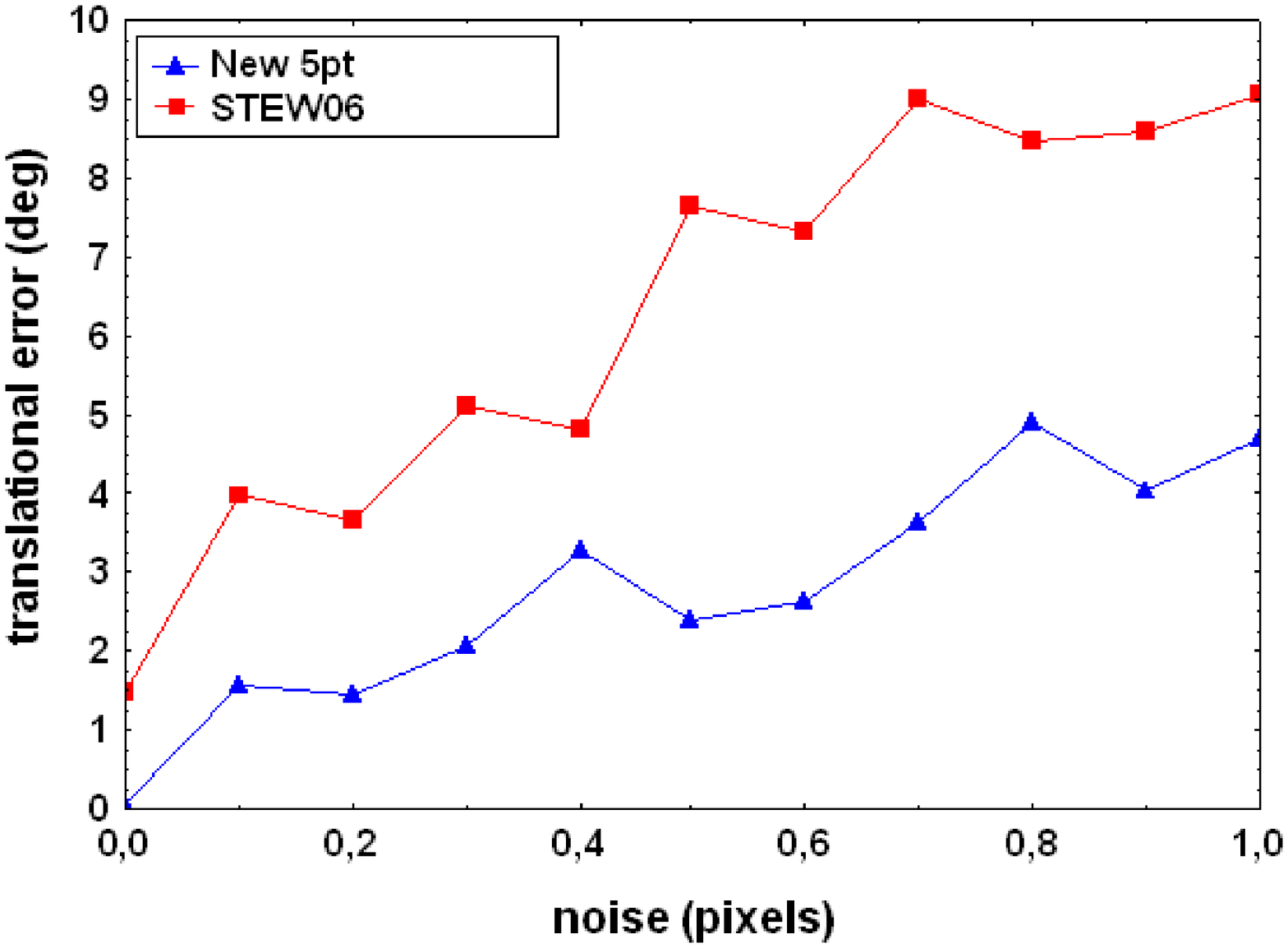}\\
      (a) & (b)\\
    \end{tabular}
    \caption{Error on the base orientation (in degree). Easy Case, a) sideway motion. b) forward motion. }
\end{figure}
\subsubsection{Planar Structure and short base}.
Several surfaces are known as "dangerous" \cite{PhilipCritical} the reason of this appellation is due to the fact that if the points chosen for the evaluation of the relative orientation are on this kind of surface, the configuration becomes degenerate. In the following, one of the most unfavorable configurations has been chosen. We note that the method of the 5 points of Stewenius is not robust in the sideways motion case. Besides, Sarkis \cite{PlanarProblemNister} has shown this weakness of the algorithm, and concluded that it is better in such cases to use an homography. On the other hand, with the method presented here this kind of configuration does not lead to any problem.
\begin{figure}[!h]
    \centering
    \begin{tabular}{ccc}
      \includegraphics[width=0.35\linewidth]{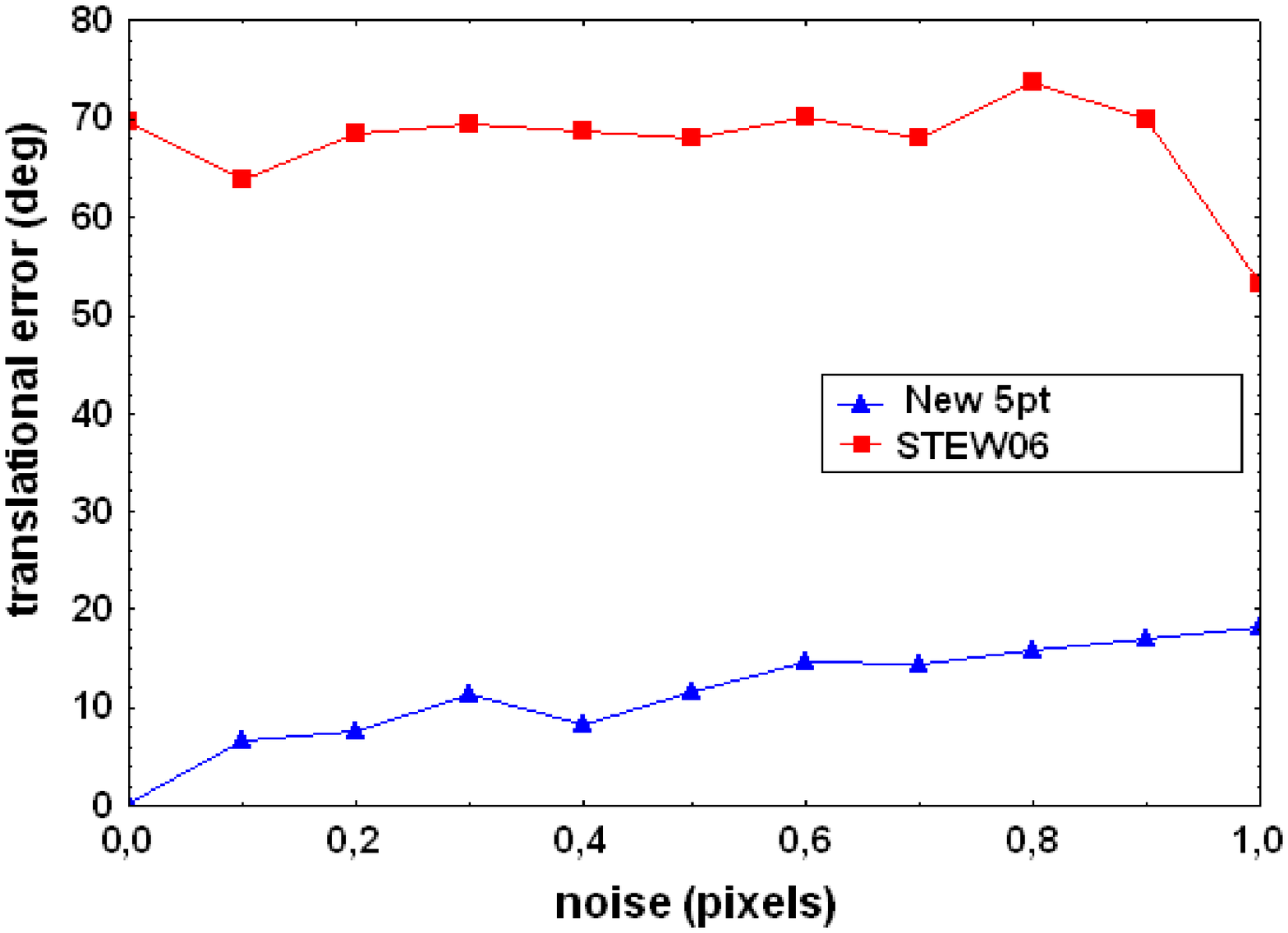}&
      \includegraphics[width=0.35\linewidth]{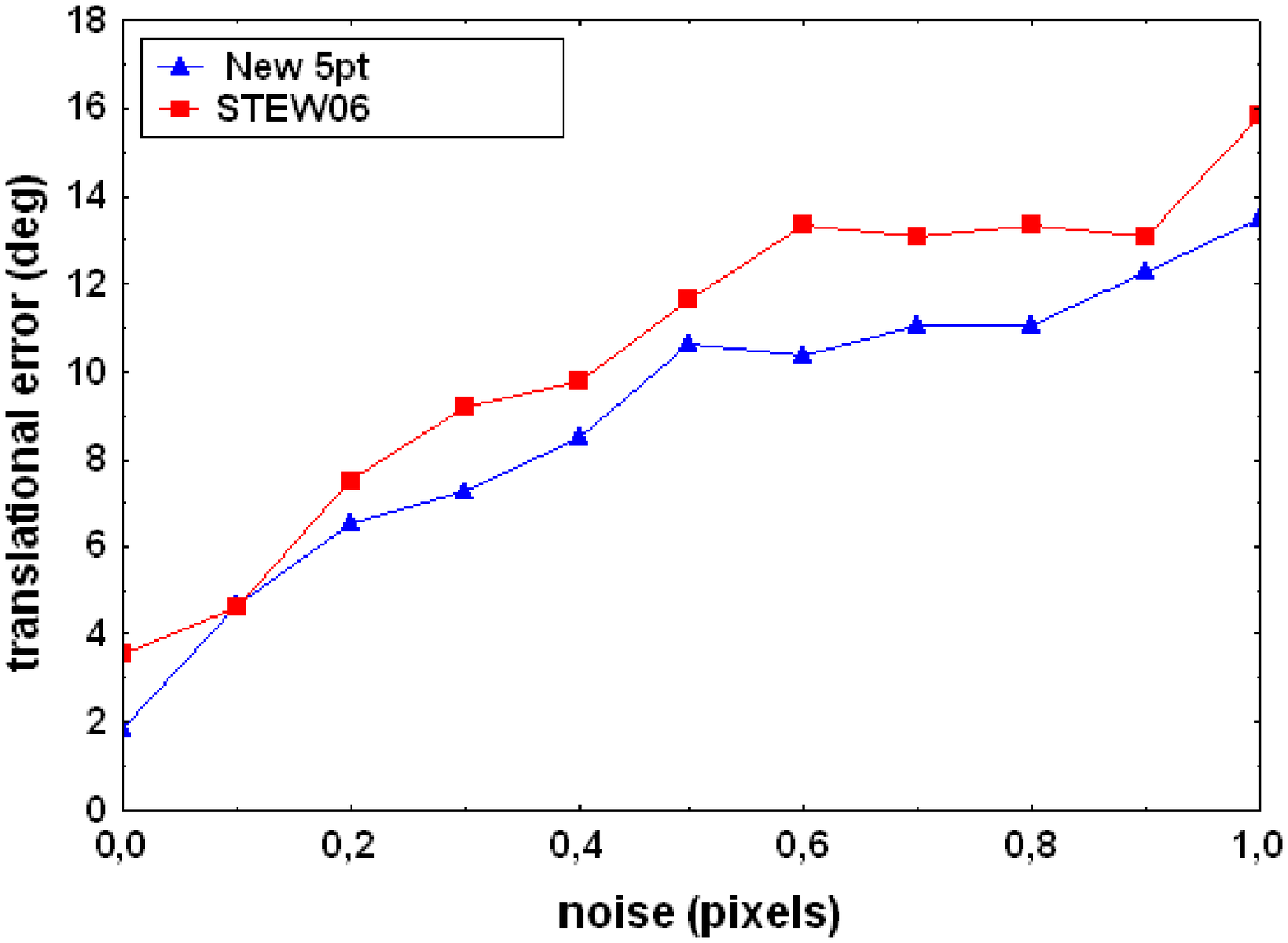}&
        \includegraphics[width=0.35\linewidth]{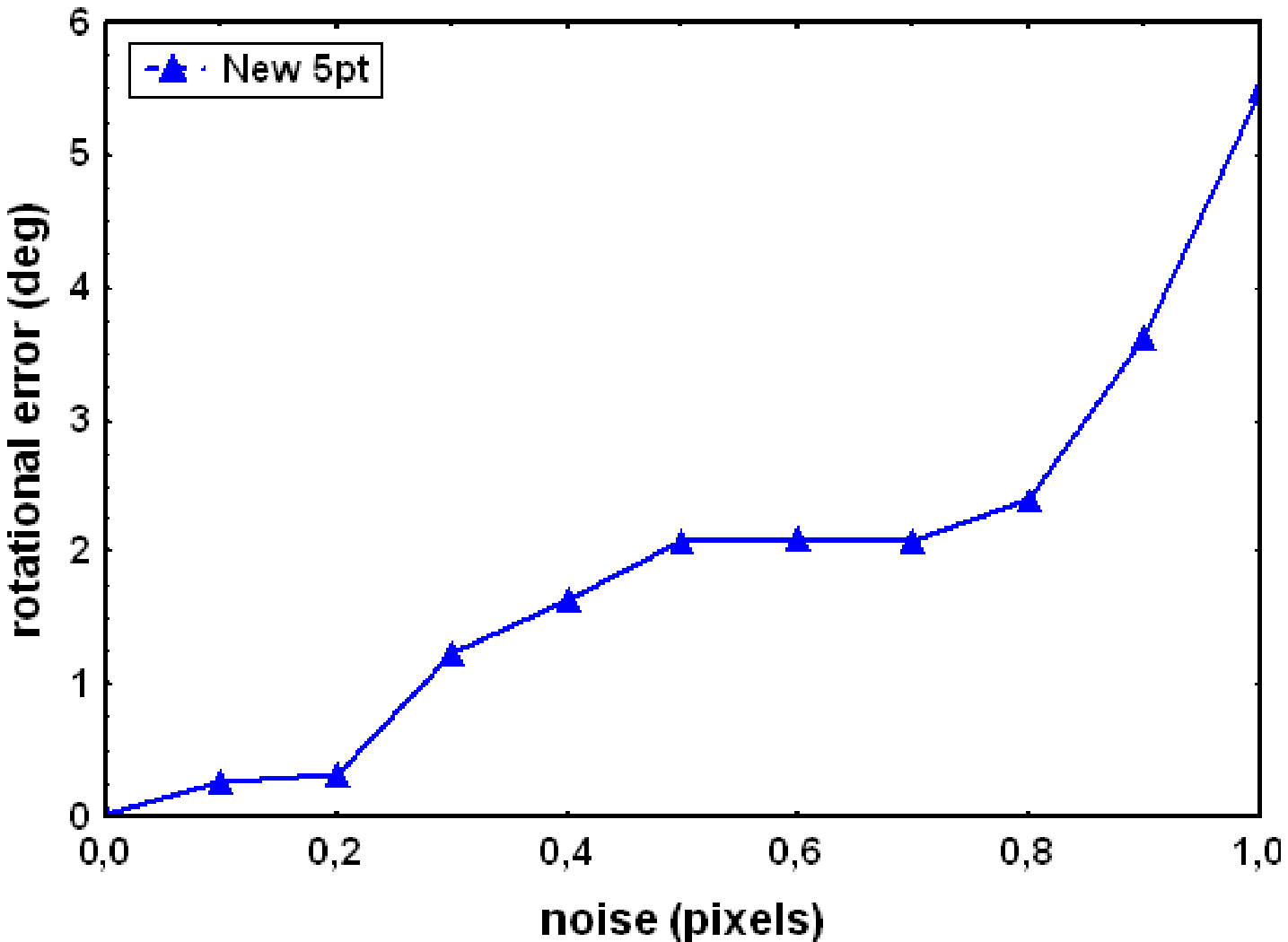}\\
      (a) & (b) & (c)\\
    \end{tabular}
    \caption{a) Error on the base orientation (in degree) and standard deviation, planar case, sideway motion b) idem, for a forward motion. c) Error on the relative orientation, non planar case, null base length.}
\label{fig:resultatsdesSimu}
\end{figure}
\subsubsection{Results for a null base length}
Even with a null translation, the rotation is very well defined. This is probably due to the fact that the parameters to estimate in our initial equations are completely separated. The rotation is not mixed with the translation. This may explain why a translation of zero length does not affect the results. Fig. \ref{fig:resultatsdesSimu}(c).
\subsection {Tests on Real Images}
So as to test the algorithm presented in section 6, we have used the recently available image base set up by ISPRS \cite{ISPRS} for relative orientation tests. We have checked especially our ambiguity resolution so as to find the good physical solution. The mean error on the baseline orientation is equal to  $5.25^\circ$ and  mean error on the rotation is $1.26^\circ$.
At the present time, an average of 20 seconds on a 1.65 GHz machine is necessary to get the solution. This time of calculation may be to improved by using offline resolutions as done by Stewenius.

\section{Conclusion}
In this paper a new method for the problem of the "five points pose problem"
has been described. The main difference with the previous methods is that the rotation and the translation are directly the unknowns of the system. Then we have shown that with the available tools of geometric algebra, this kind of system can be solved. The major advantage, when compared to other methods, is that it works accurately on all cases of plane scenes, or on couples of images with a null base.

\bibliographystyle{splncs}

\end{document}